\documentclass[letterpaper]{article} 
\PassOptionsToPackage{table}{xcolor} 
\usepackage{aaai2026}  
\usepackage{times}  
\usepackage{helvet}  
\usepackage{courier}  
\usepackage[hyphens]{url}  
\usepackage{graphicx} 
\usepackage{natbib}  
\usepackage{caption} 
\usepackage{algorithm}
\usepackage{algorithmic}

\usepackage{newfloat}
\usepackage{listings}
\DeclareCaptionStyle{ruled}{labelfont=normalfont,labelsep=colon,strut=off} 
\floatstyle{ruled}
\newfloat{listing}{tb}{lst}{}
\floatname{listing}{Listing}
\usepackage{booktabs} 
\usepackage[most]{tcolorbox} 
\usepackage{amssymb} 
\usepackage{tabularx} 
\usepackage{float}

\definecolor{mycolor}{rgb}{0.4, 0.7, 0.8}
\definecolor{deepgreen}{RGB}{0, 128, 0}

\newtcolorbox{mybox}[2][]
  {colback = black!5!white, colframe = black!75!black, fonttitle = \bfseries,
    colbacktitle = black!100!black, enhanced,
    attach boxed title to top left={yshift=-2.2mm,xshift=4mm},
    title=#2,#1}

\title{MedGen: Unlocking Medical Video Generation \\ by Scaling Granularly-annotated Medical Videos}
\author{
    Rongsheng Wang\textsuperscript{\rm 1}\equalcontrib
    Junying Chen\textsuperscript{\rm 1}\equalcontrib,
    Ke Ji\textsuperscript{\rm 1},\\
    Zhenyang Cai\textsuperscript{\rm 1},
    Shunian Chen\textsuperscript{\rm 1},
    Yunjin Yang\textsuperscript{\rm 1},
    Benyou Wang\textsuperscript{\rm 1}\thanks{Corresponding author.}
}
\affiliations{
    \textsuperscript{\rm 1}The Chinese University of Hong Kong, Shenzhen\\
    \textit{wangbenyou@cuhk.edu.cn}\\
    {\large \color{magenta}{\url{https://github.com/FreedomIntelligence/MedGen}}}

}

\usepackage{bibentry}

\begin{document}

\maketitle

\begin{abstract}
Recent advances in video generation have shown remarkable progress in open-domain settings, yet medical video generation remains largely underexplored. Medical videos are critical for applications such as clinical training, education, and simulation, requiring not only high visual fidelity but also strict medical accuracy. However, current models often produce unrealistic or erroneous content when applied to medical prompts, largely due to the lack of large-scale, high-quality datasets tailored to the medical domain.
To address this gap, we introduce \textbf{MedVideoCap-55K}, the first large-scale, diverse, and caption-rich dataset for medical video generation. It comprises over 55,000 curated clips spanning real-world medical scenarios, providing a strong foundation for training generalist medical video generation models. Built upon this dataset, we develop \textbf{MedGen}, which achieves leading performance among open-source models and rivals commercial systems across multiple benchmarks in both visual quality and medical accuracy.
We hope our dataset and model can serve as a valuable resource and help catalyze further research in medical video generation.

{\color{mycolor}\textbf{\includegraphics[height=1em]{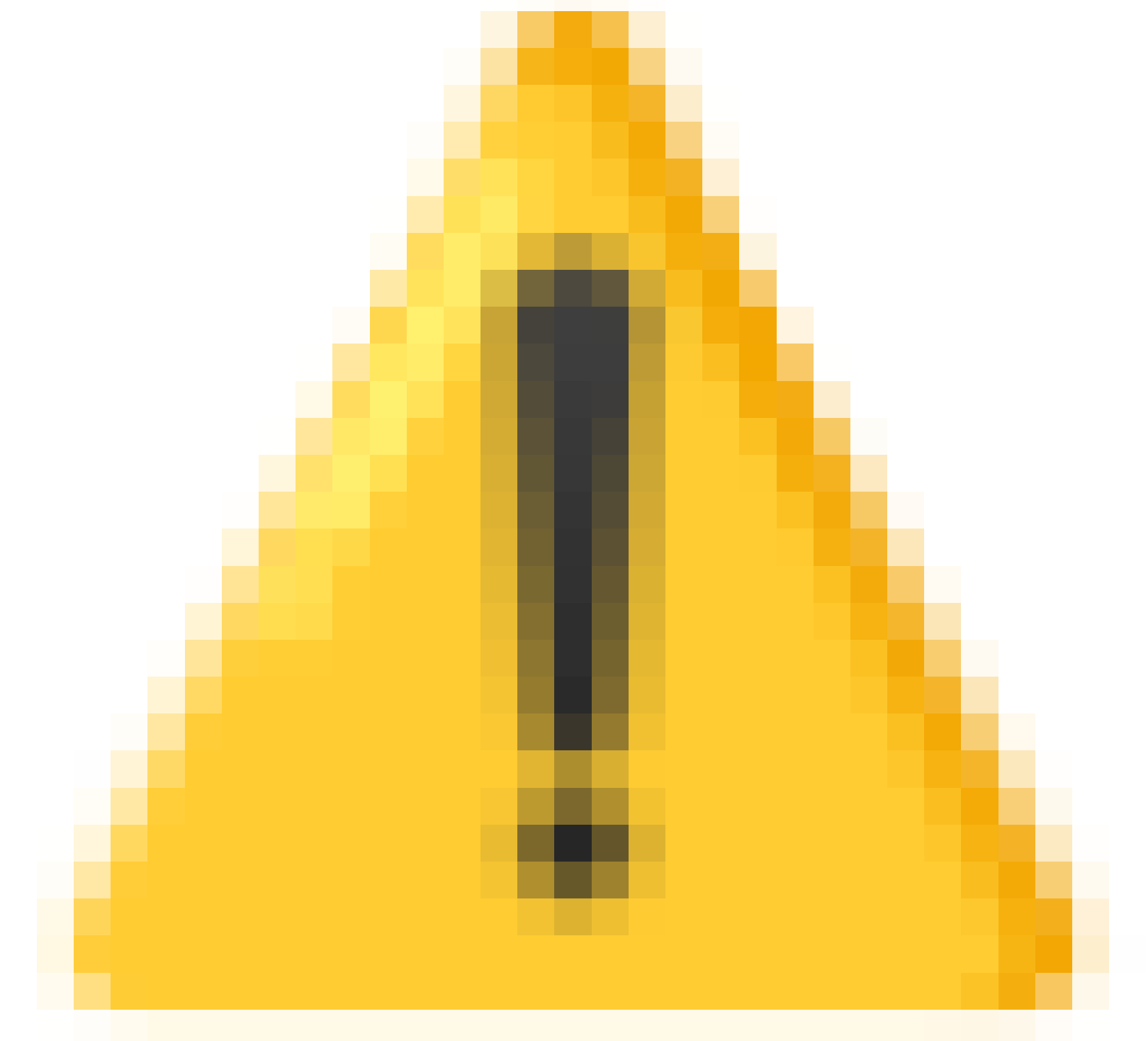} Disclaimer: This paper contains clinical content that may be disturbing to some readers.}}
\end{abstract}

\section{Introduction}

Recent advances in video generation have led to impressive breakthroughs, with models now capable of producing high-quality, cinematic visuals that align closely with user prompts~\cite{blattmann2023stable}. In particular, latent video diffusion models (LVDMs), such as Sora~\cite{sora} and Veo~\cite{veo2024}, have achieved state-of-the-art performance by operating efficiently in latent space and delivering diverse, coherent video outputs from textual descriptions.

Despite this progress, \textbf{medical video generation} remains a largely underexplored yet crucial domain. Medical videos are indispensable in numerous real-world applications, including clinical training, surgical simulation, and patient education~\cite{li2024artificialintelligencebiomedicalvideo}. Unlike everyday video content, medical videos demand precise rendering of anatomical structures, accurate surgical steps, and realistic physiological dynamics. These requirements place significantly higher demands on visual fidelity, semantic correctness, and temporal coherence.

\begin{figure}[ht!]
    \centering
    \includegraphics[width=0.40\textwidth, trim={0mm 13mm 0mm 0mm}, clip]{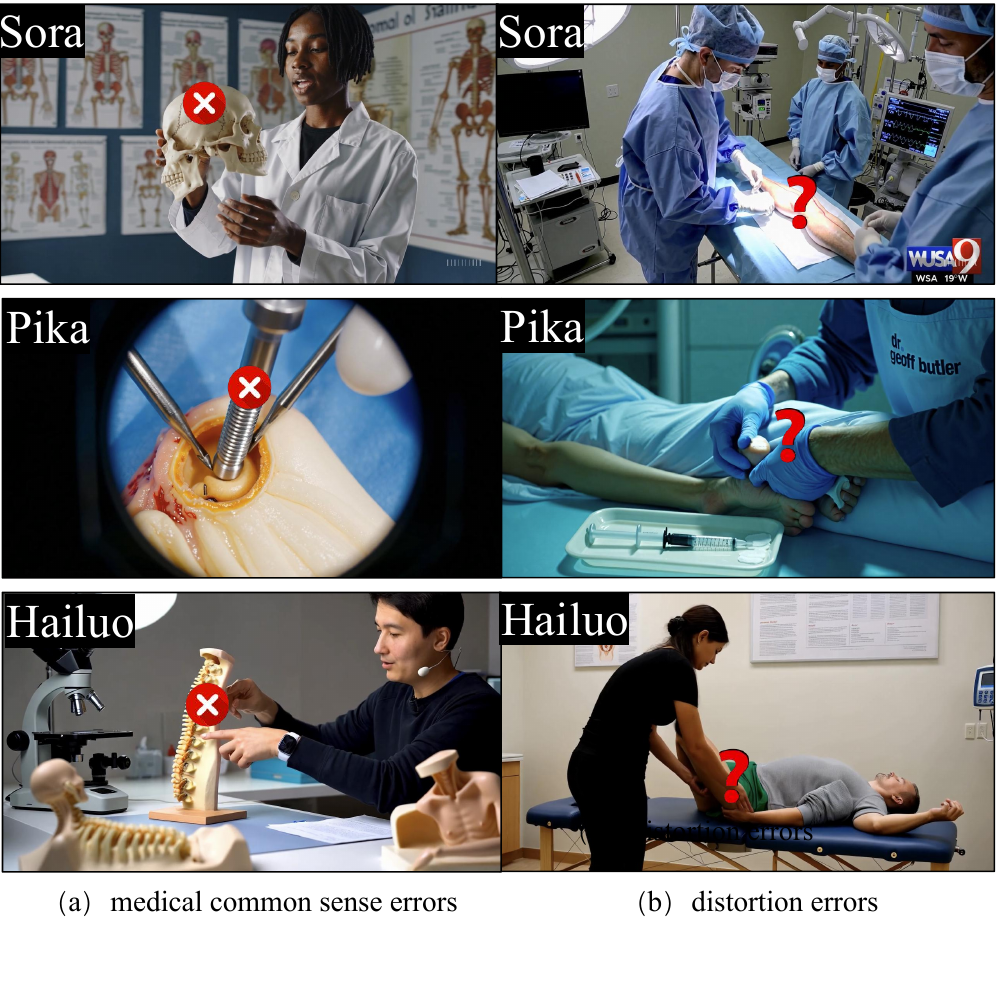}
    \caption{Failure cases of Sora, Pika, and Hailuo on medical video generation. (a): Medical common sense errors. (b): Distortion errors.}
    \label{fig:pilot-study}
\end{figure}

\begin{table*}[t!]
\footnotesize
\caption{Comparison of existing medical video datasets. MedVideoCap-55K offers the largest scale, highest diversity, and is the first to include detailed text captions tailored for medical text-to-video generation. “\#” and “*” denote “number” and “average.” \includegraphics[height=1em]{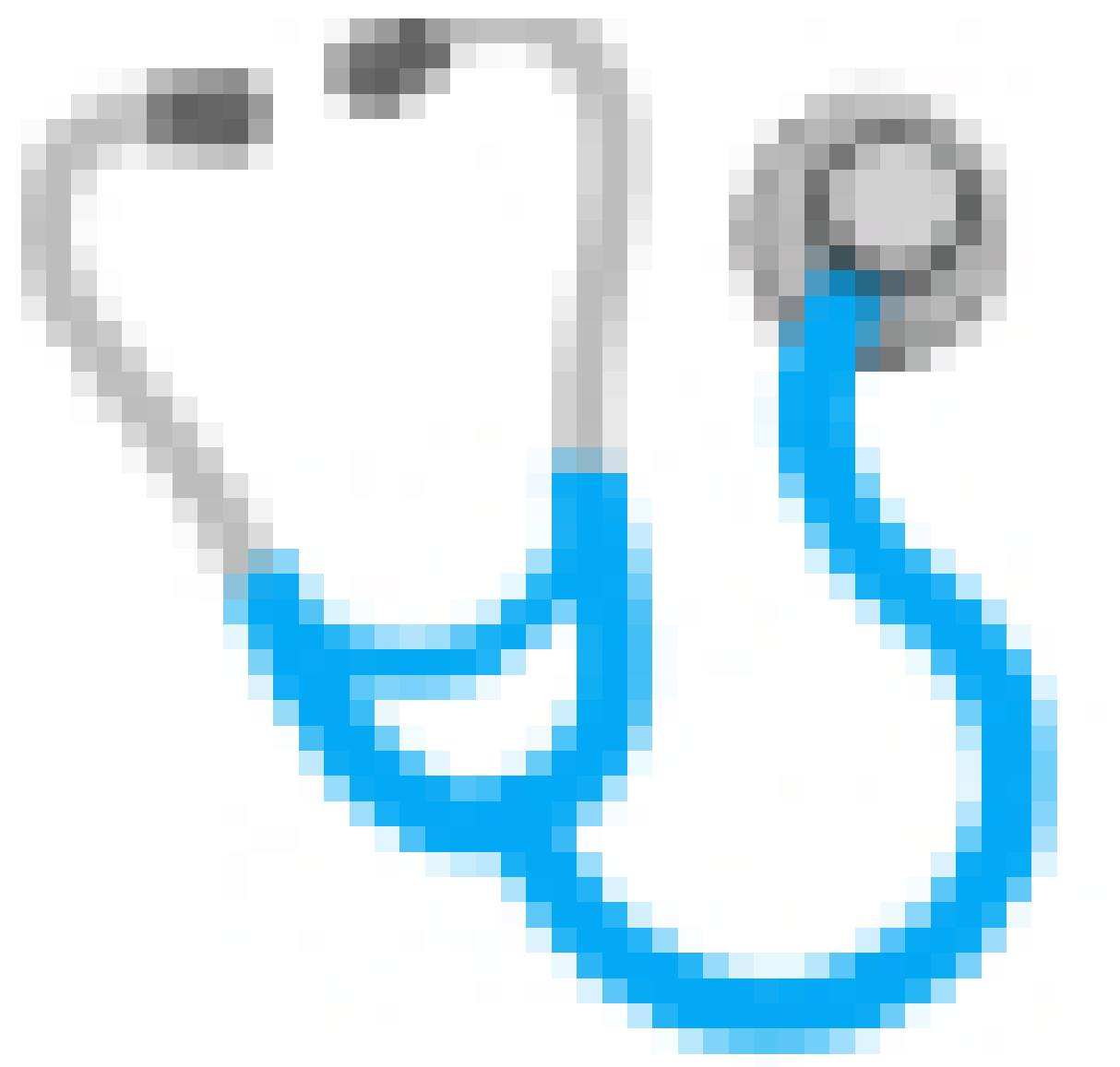}: clinical practice, \includegraphics[height=1em]{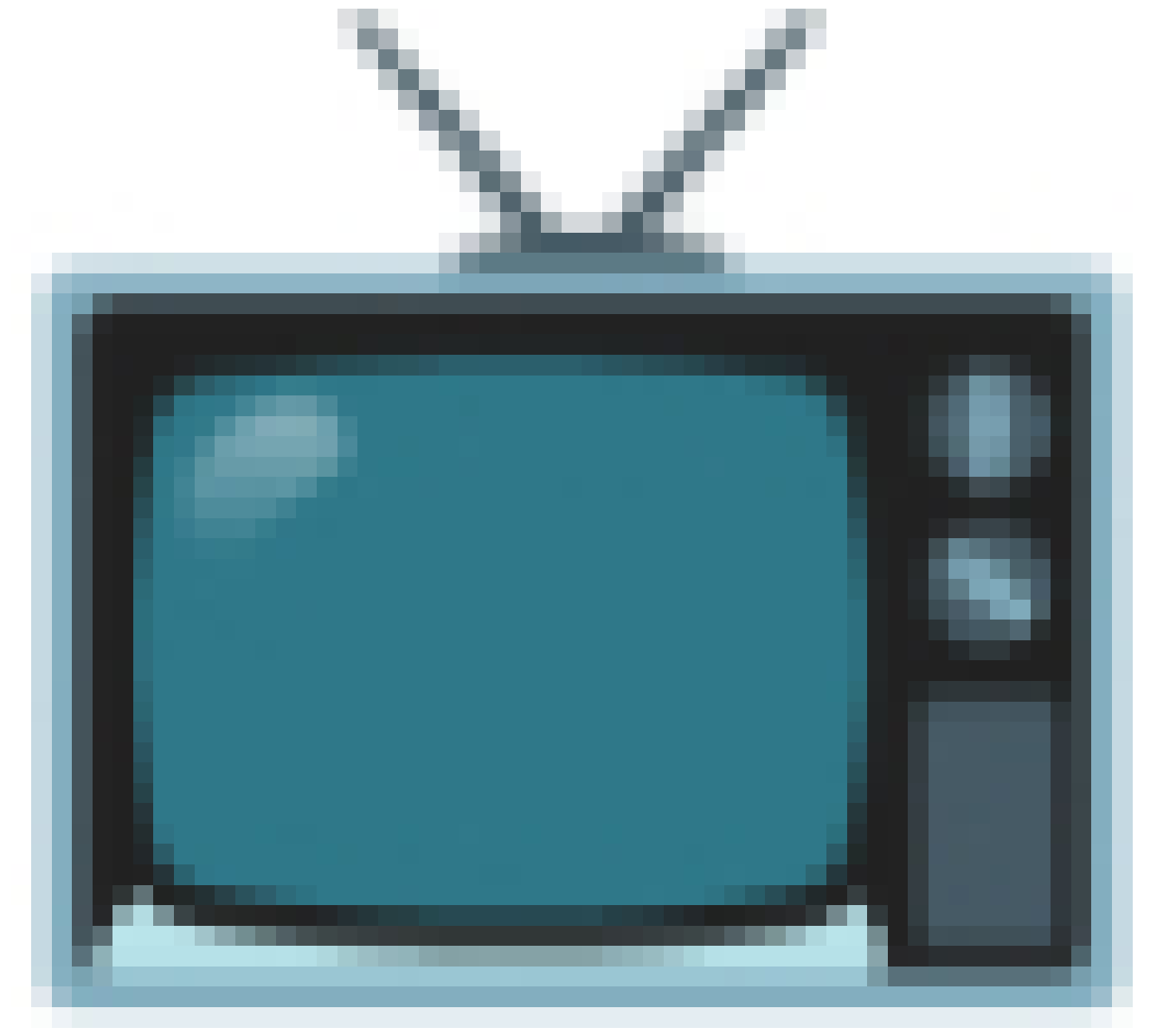}: medical science, \includegraphics[height=1em]{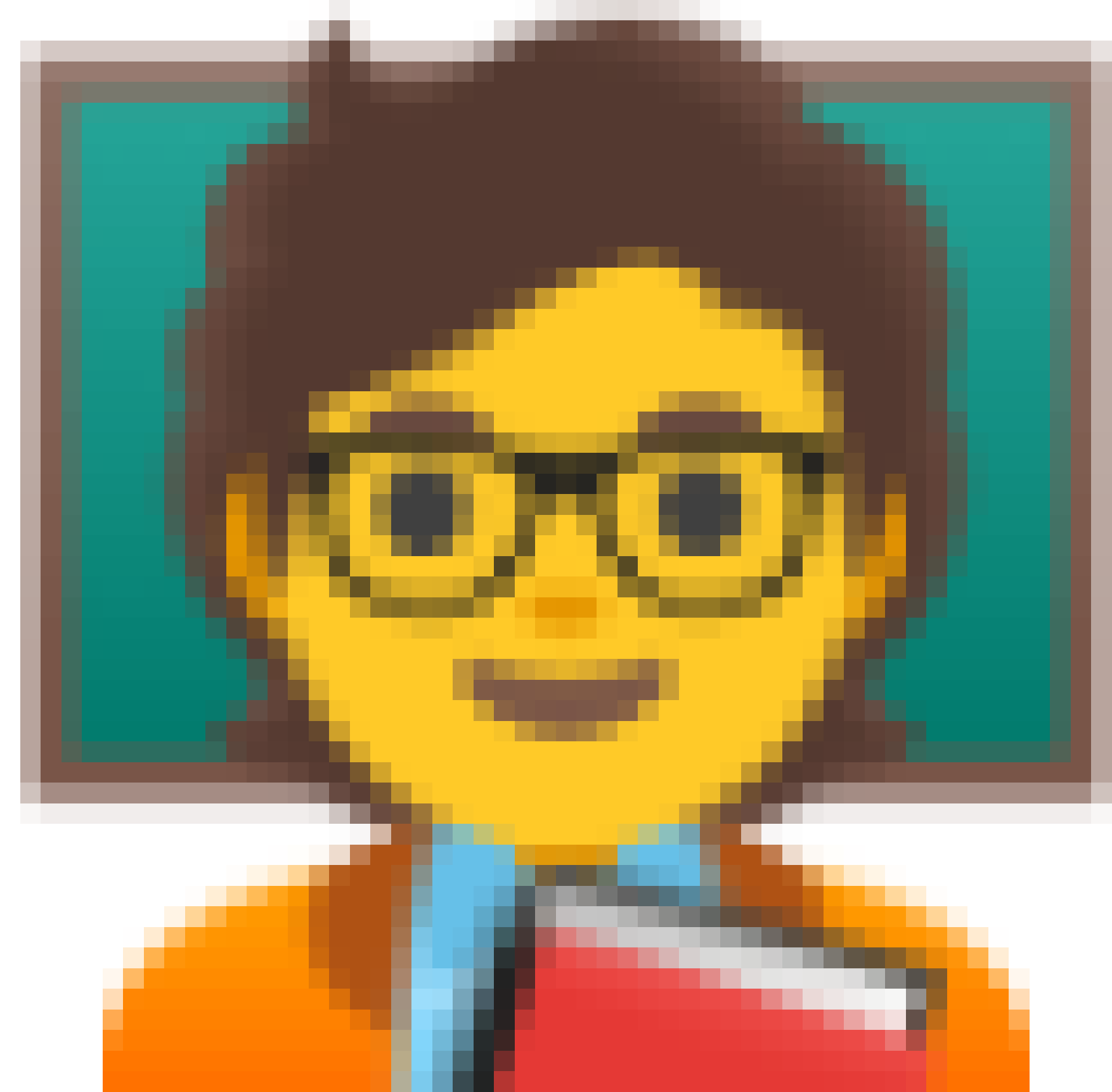}: medical teaching, \includegraphics[height=1em]{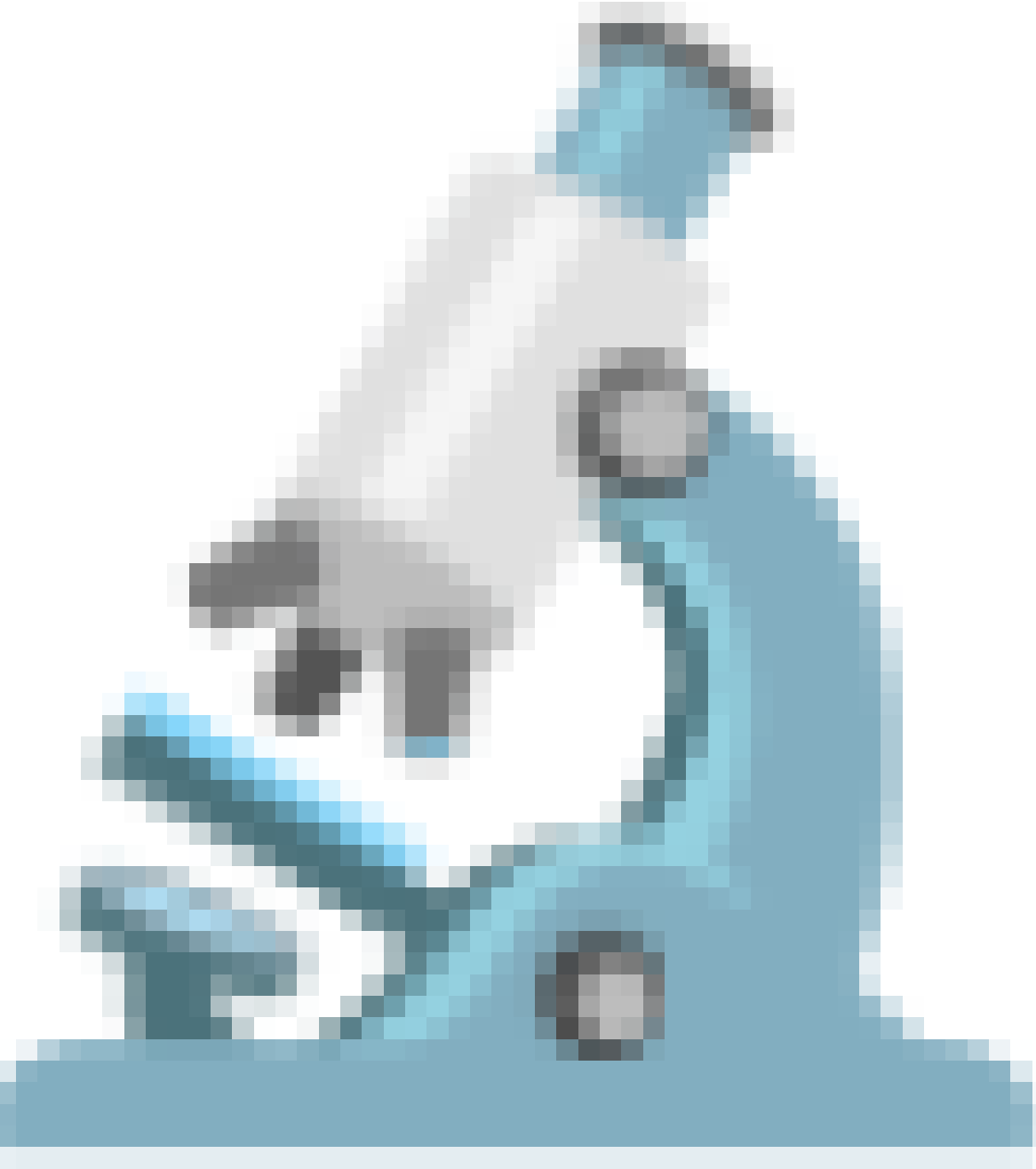}: medical imaging, \includegraphics[height=1em]{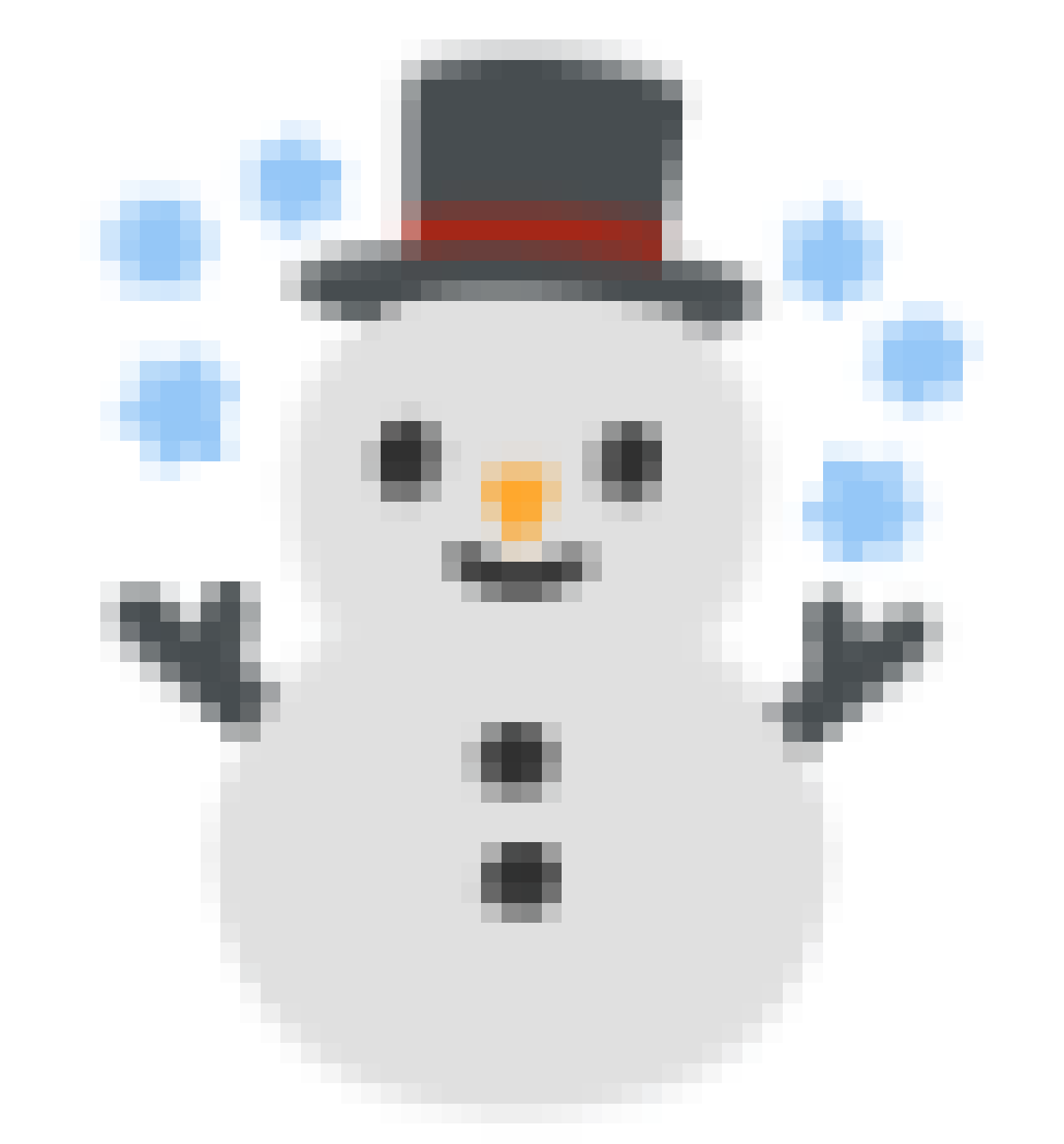}: medical animation.}
\centering
\resizebox{0.9\textwidth}{!}{
\begin{tabular}{llllcll}
\hline
Dataset & Caption & \#Vid. & Len.* & Words* & Resolution & Category \\
\hline
HyperKvasir \cite{borgli2020hyperkvasir} & Category Labels & 120 & 4s & None & 512 × 320 & \includegraphics[height=1em]{figures/icon_clinical_practice.pdf} \\
SurgicalActions160 \cite{schoeffmann2018video}  & Category Labels & 160 & 2s & None &  320 × 240 & \includegraphics[height=1em]{figures/icon_clinical_practice.pdf} \\
Pitvis \cite{das2024pitvis} & Category Labels & 25 & 62m & None & 1280 × 720 & \includegraphics[height=1em]{figures/icon_clinical_practice.pdf} \\
CatRelDet \cite{ghamsarian2021relevance} & Category Labels & 2,200 & 3s & None & 224 × 224 & \includegraphics[height=1em]{figures/icon_clinical_practice.pdf}  \\
MedvidCL  \cite{gupta2023dataset} & Category Labels & 6,117 & 4s & None & 512 × 320 & \includegraphics[height=1em]{figures/icon-teacher.pdf} \\
Cholec80 \cite{twinanda2016endonet} & Category Labels & 80 & 50m & None & 1920 × 1080  & \includegraphics[height=1em]{figures/icon_clinical_practice.pdf}  \\
Colonoscopic \cite{mesejo2016computer} & Category Labels &  76 & 20s & None & 768 × 576  & \includegraphics[height=1em]{figures/icon_clinical_practice.pdf} \\
MESAD-Real \cite{bawa2021saras} & Category Labels & 23,366 & 5s & None & 720 × 756 & \includegraphics[height=1em]{figures/icon_clinical_practice.pdf} \\
SurgToolLoc \cite{zia2023surgical} & Category Labels & 24,695 & 30s & None & 1280 × 720 & \includegraphics[height=1em]{figures/icon_clinical_practice.pdf} \includegraphics[height=1em]{figures/icon-teacher.pdf} \\
Bora \cite{sun2024bora} & Text Captions &  4,897 & 5s & 110 & 256 × 256  & \includegraphics[height=1em]{figures/icon_clinical_practice.pdf} \\
\hline
\rowcolor{orange!5} \textbf{MedVideoCap-55K (ours)} & Text Captions &  \textbf{55,803} &  8s &  174 & 720 × 480 & \includegraphics[height=1em]{figures/icon_clinical_practice.pdf} \includegraphics[height=1em]{figures/icon-television.pdf} \includegraphics[height=1em]{figures/icon-teacher.pdf}  \includegraphics[height=1em]{figures/icon-microscope.pdf} \includegraphics[height=1em]{figures/icon-snowman.pdf} \\
\hline
\end{tabular}
}
\label{tab:compare_data}
\end{table*}

However, current video generation models are trained almost exclusively on general-purpose datasets that focus on natural scenes and everyday activities~\cite{blattmann2023stable}. As a result, when applied to medical prompts, they often generate outputs with critical errors—such as anatomical inconsistencies, tool misuse, and implausible clinical scenarios. As shown in Figure~\ref{fig:pilot-study}, even leading models like Sora~\cite{sora}, Pika~\cite{pika}, and Hailuo~\cite{hailuo} fail to maintain basic medical realism, revealing a clear mismatch between training data and medical domain requirements. A key bottleneck lies in the lack of large-scale, high-quality datasets tailored for medical video generation~\cite{sun2024bora}. Existing medical datasets are limited in size, narrow in scope (e.g., only endoscopic or surgical videos), and mostly provide categorical labels instead of detailed descriptions—making them unsuitable for training or evaluating text-to-video generation models that rely on rich, semantic input.

To address this gap, we introduce \textbf{MedVideoCap-55K}, the first large-scale medical video dataset specifically designed for text-to-video generation. It contains over 55K carefully filtered video clips spanning a wide spectrum of real-world medical scenarios, including clinical practice, imaging, education, animation, and science popularization. Each video is paired with high-quality, descriptive captions generated by multimodal large language models (MLLMs). The dataset is curated using a rigorous filtering pipeline that ensures high visual quality, medical relevance, and training suitability.

Building on MedVideoCap-55K, we develop \textbf{MedGen}, a specialized medical video generation model. Through comprehensive experiments, we demonstrate that MedGen significantly outperforms other open-source models in both visual quality and medical accuracy. In addition, we propose evaluation metrics tailored to medical video generation—addressing gaps in current benchmarks—and showcase downstream applications such as medical data augmentation and simulation.

Our key contributions are summarized as follows:

\begin{enumerate}
  \item[1.] We introduce \textbf{MedVideoCap-55K}, the first large-scale medical video dataset designed for text-to-video generation, containing 55K diverse, high-quality medical clips with detailed captions.
  \item[2.]  We develop \textbf{MedGen}, a medical video generation model trained on MedVideoCap-55K. MedGen excels in medical video generation, demonstrating strong performance in both visual quality and medical accuracy.
  \item[3.] We propose a suite of evaluation protocols tailored for medical video generation, including both automatic metrics and expert-based assessments, addressing the limitations of existing general-purpose benchmarks.
\end{enumerate}

\section{Motivations}

\textbf{Existing Models Fall Short  in Medical Video Generation}~
Despite recent advances in general video generation, existing commercial models such as Sora~\cite{sora}, Pika~\cite{pika}, and Hailuo~\cite{hailuo} struggle in the medical domain. As shown in Figure~\ref{fig:pilot-study}, these models frequently generate content with severe medical hallucinations—such as incorrect anatomical structures, misuse of instruments, or implausible clinical scenes. These failures are not isolated cases but a systemic issue rooted in a lack of medical-specific knowledge and training data. Current models are optimized for general visual patterns, not the precise semantics, constraints, and domain logic required in medical content.

\textbf{Challenges: Limited  Medical Video Datasets}~
A core reason for these failures is the absence of high-quality, large-scale medical video datasets. As summarized in Table~\ref{tab:compare_data}, existing datasets are limited in scope—often focused on narrow tasks such as surgical action recognition or endoscopy classification—and rarely include descriptive captions. This makes them insufficient for training or evaluating text-to-video generation models that require semantic understanding and rich supervision.

To bridge this gap, we introduce \textbf{MedVideoCap-55K}, the first large-scale, caption-rich dataset specifically designed for medical video generation. Sourced from over 25 million public videos, MedVideoCap-55K contains more than 55K curated medical video clips covering diverse domains such as clinical practice, education, imaging, and animation. Each video is paired with high-quality textual descriptions, enabling training and evaluation in natural language-driven generation tasks. We believe MedVideoCap-55K provides the foundation needed to advance faithful, diverse, and clinically meaningful medical video generation.

\section{Scaling Granularly-annotated Medical Videos}
\label{sec:data}

\begin{figure*}[ht]
    \centering
    \includegraphics[width=0.98\textwidth]{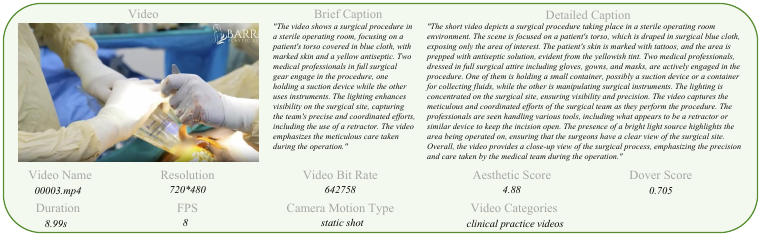}
    \caption{Sample from MedVideoCap-55K. Each data point consists of a medical video clip, a brief caption, and a detailed caption.}
    \label{fig:data_info}
\end{figure*}

In this section, we aim to scale up a high-quality medical video dataset, resulting in MedVideoCap-55K. As illustrated in Figure~\ref{fig:data_info}, each sample consists of a medical video clip paired with a granularly-annotated textual caption. Next, we will respectively cover dataset construction, enhanced data filtering, and dataset statistics.

\subsection{Data Construction}
\label{sec_31}

\begin{figure}[ht!]
    \centering
    \includegraphics[width=0.45\textwidth]{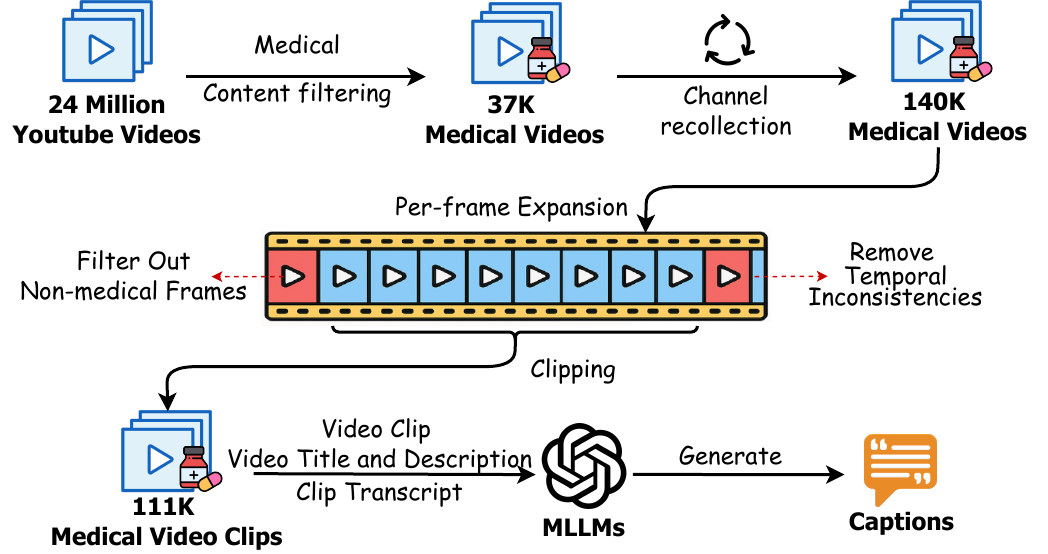}
    \caption{Overview of the data construction process.}
    \label{fig:dataset_fig1}
\end{figure}

\textbf{Collecting Videos from YouTube}

We construct MedVideoCap-55K using publicly available videos from YouTube.
To ensure scalability, we began with a large-scale collection of 25 million YouTube videos along with their associated titles and descriptions. We applied a two-stage filtering pipeline to identify medically relevant content. First, we used a curated medical keyword dictionary to flag potentially relevant videos. Second, we applied a text classifier trained on \textit{snowflake-arctic-embed-m}~\cite{merrick2024arctic} to further verify medical relevance based on video metadata. This yields 37K medical videos. To expand coverage, we collect additional videos from the channels that hosted these medical videos, ultimately assembling a total of \textbf{140K medical videos} with a total duration of \textbf{10,269 hours}.

\textbf{Video Segmentation}
To extract coherent segments rich in medical content, we applied the following video segmentation methods:

\begin{enumerate}
    \item \textbf{Medical Frame Classification:} Leveraging the visual encoder from \textit{CLIP} \cite{clip} and human-annotated data, we trained a frame-level classifier $C$. Videos were sampled at 1 frame per second (FPS), and each frame \( x_i \) (the i-th frame) was labeled as either medical or non-medical: \( C(x_i) \in \{1, 0\} \).
    
    \item \textbf{Temporal Consistency:} To ensure visual coherence, we calculated similarity scores \( S(x_i, x_{i-1}) \) between adjacent frames using CLIP embeddings. Only consecutive frames with similarity above a threshold \( \tau \) were retained.
\end{enumerate}

A valid clip \( [x_i, x_j] \) was preserved if it satisfied the following conditions:
$$
\mathbb{I}[j - i \geq 6] \cdot \prod_{k=i}^{j} \mathbb{I}[C(x_k) = 1] \cdot \prod_{k=i+1}^{j} \mathbb{I}[S(x_k, x_{k-1}) > \tau] = 1
$$

Finally, we kept only clips with a resolution above 480p, resulting in  \textbf{111K medical video clips}.  The overall data construction pipeline is illustrated in Figure~\ref{fig:dataset_fig1}.

\textbf{Caption Generation}
We leverage a multimodal LLM (GPT-4o) to generate detailed captions. Due to context limits, we uniformly sampled 8 frames per clip as visual input. To minimize hallucinations, we supply the video title and description, along with the transcript of the clip. To support models with input length constraints, we further prompt the MLLM to generate a brief caption.

{ \small
\begin{mybox}[colback=gray!10]{Prompt MLLM to Generate Video Caption}

The provided images are sampled from a medical video clip (8 evenly spaced frames). This clip is taken from a video with the following metadata:  

\textbf{Video Title:} \textcolor{mycolor}{\{Video Title\}}

\textbf{Video Description:} \textcolor{mycolor}{\{Video Description\}}

\textbf{The transcript of this medical clip is as follows:} 

\textcolor{mycolor}{\{Clip Transcript\}}

Using the visual content of the clip, along with the title, description, and transcript, please generate a clear, detailed, and accurate description of what is shown in this clip. Focus on visible scenes, actions, and medical elements, including any procedures, instruments, or clinical interactions shown.

\end{mybox}
}

\subsection{Data Filtering and  Quality Refinement}
\label{sec_32}

\begin{figure*}[t]
    \centering
    \includegraphics[width=\textwidth]{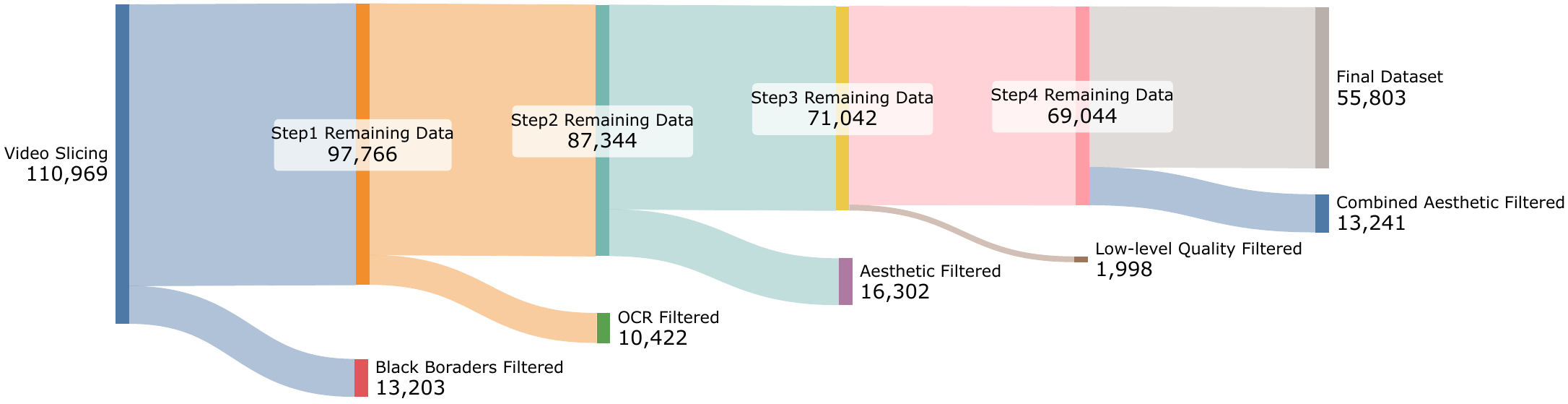}
    \caption{Overview of our data filtering pipeline. Each stage applies specific filters and shows the volume of data removed and retained.}
    \label{fig:filter_pipeline}
\end{figure*}

\begin{figure*}[t]
    \centering
    \begin{minipage}[c]{0.50\textwidth}
        \centering
        \includegraphics[width=\textwidth]{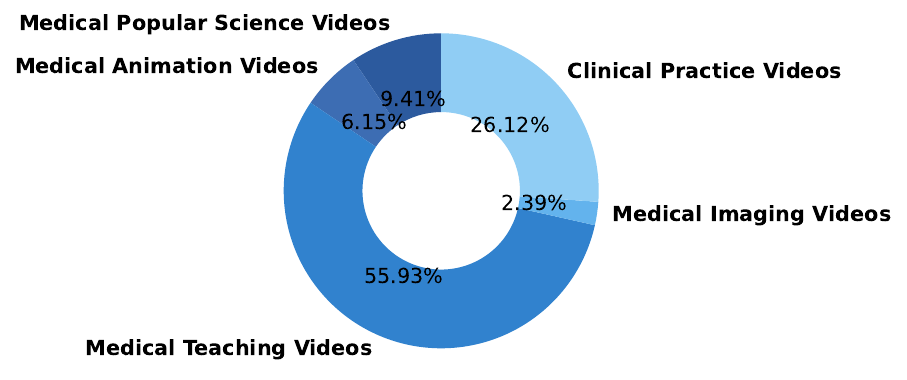}
        \caption*{(a) Category Distribution}
    \end{minipage}
    \hfill
    \begin{minipage}[c]{0.48\textwidth}
        \centering
        \begin{minipage}[c]{0.48\textwidth}
            \centering
            \includegraphics[width=\textwidth]{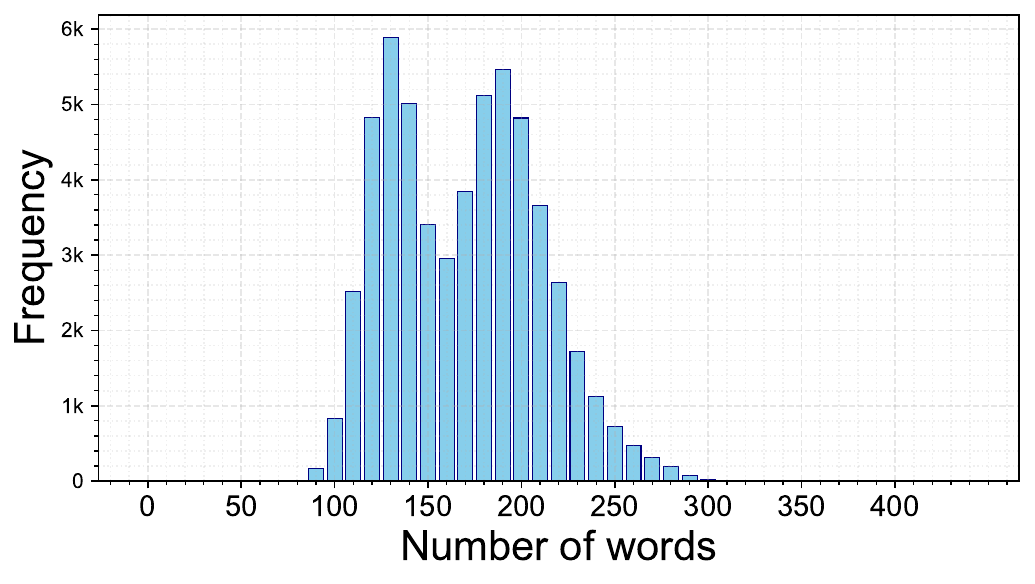}
            \caption*{(b) Detailed Captions}
        \end{minipage}
        \hfill
        \begin{minipage}[c]{0.48\textwidth}
            \centering
            \includegraphics[width=\textwidth]{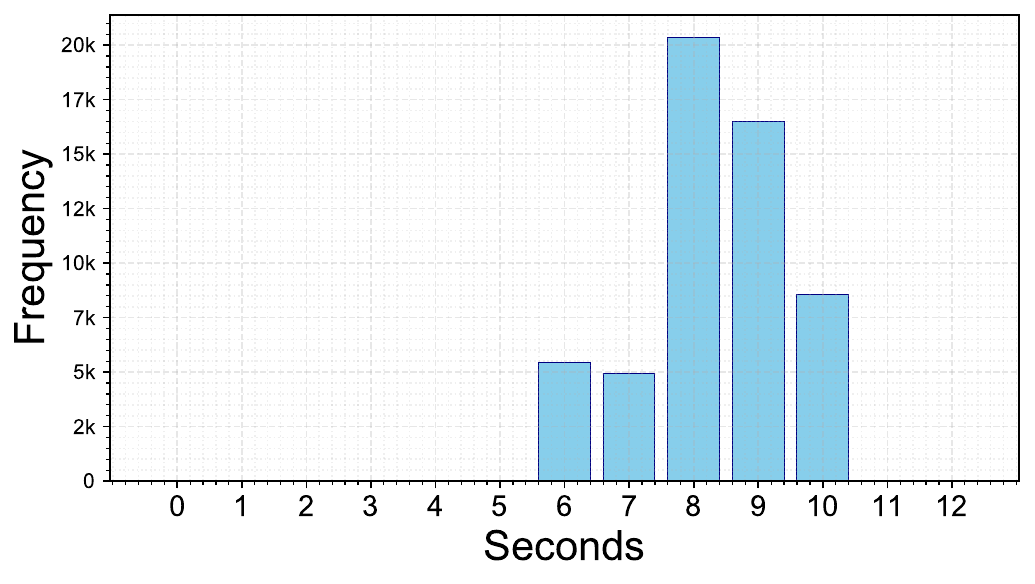}
            \caption*{(c) Duration}
        \end{minipage}
        \begin{minipage}[c]{0.48\textwidth}
            \centering
            \includegraphics[width=\textwidth]{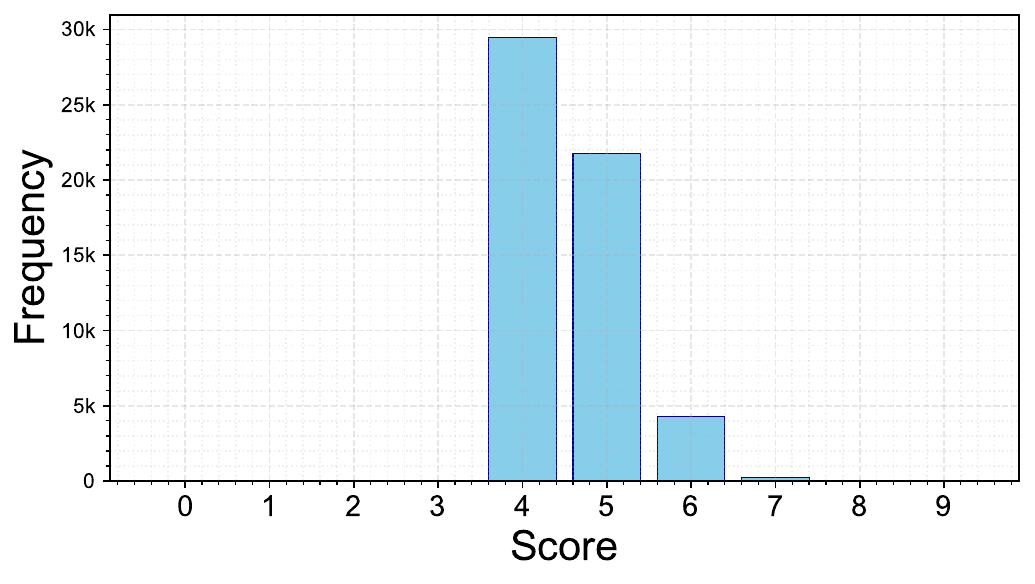}
            \caption*{(d) Aesthetic Quality}
        \end{minipage}
        \hfill
        \begin{minipage}[c]{0.48\textwidth}
            \centering
            \includegraphics[width=\textwidth]{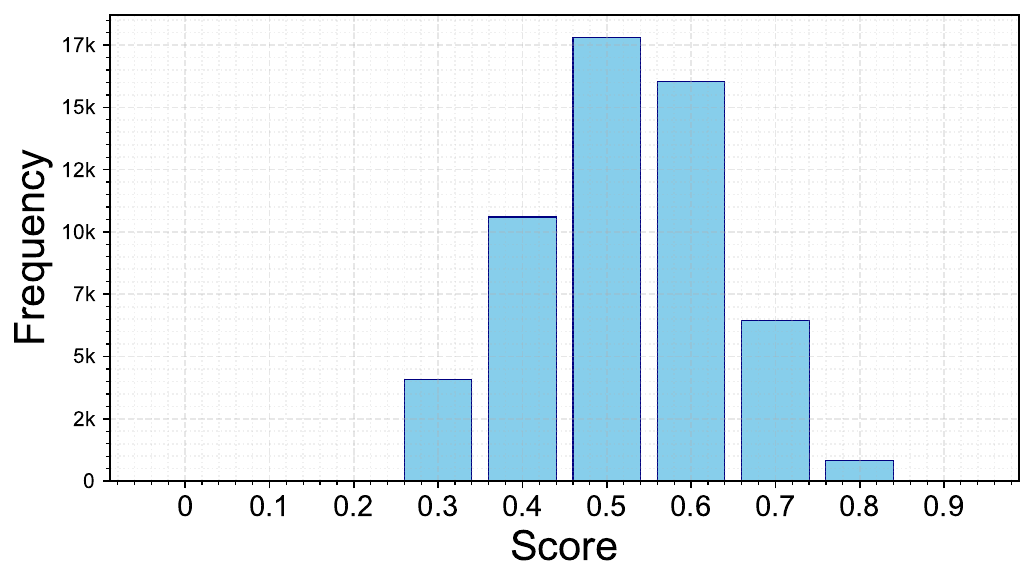}
            \caption*{(e) Dover}
        \end{minipage}
    \end{minipage}
    \caption{Data distribution in our MedVideoCap-55K. (a): category distribution of medical videos. (b): word count distribution in detailed captions. (c): duration of video clips. (d): aesthetic score distribution. (e): dover score distribution.}
    \label{fig:data_distributions}
\end{figure*}

To further improve data quality, we apply a second-stage filtration after initial segmentation.
While earlier steps remove obviously irrelevant or low-resolution videos, many subtle yet impactful quality issues remain—such as persistent black borders, heavy subtitles, visual clutter, or technical artifacts. These imperfections are difficult to detect through simple heuristics, but they can significantly degrade the learning of video generation models. To address this, we design a fine-grained filtering pipeline that systematically removes such noisy content. 
The full filtration process and the number of videos removed at each stage are visualized in Figure~\ref{fig:filter_pipeline}. 
However, each stage processes a large amount of data, making full human validation impractical. Instead, we randomly sample 200 videos at each stage and evaluate them across key quality dimensions. A stage is considered successful if at least 95\% of the sampled videos are rated as clean or only slightly noisy.

\textbf{Black Border Removal}
We use OpenCV\footnote{\url{https://github.com/opencv/opencv-python}} to detect black borders, which frequently occur in screen-recorded or lecture-style videos. These borders introduce large static regions that harm the spatial modeling capacity of generative models. Specifically, we apply Canny edge detection followed by Hough line transforms to localize consistent black edges along the video frame boundaries. Videos with persistent borders on any side are excluded.

\textbf{Subtitle Filtering via OCR}
To detect overlaid subtitles that may obscure key visual information, we apply EasyOCR\footnote{\url{https://github.com/JaidedAI/EasyOCR}} on 5 uniformly sampled frames from each video. If the total number of detected words exceeds 20, the video is discarded. This threshold effectively filters out clips with persistent subtitles or dialogue captions, which tend to dominate the visual field and distract from medically relevant content such as instruments or anatomy.

\textbf{Aesthetic Quality Filtering}
We assess aesthetic quality using the LAION aesthetic predictor~\footnote{\url{https://github.com/christophschuhmann/improved-aesthetic-predictor}}. By averaging scores over 5 sampled frames, we filter out videos that are overly blurry, low-resolution, poorly lit, or cluttered with logos and watermarks. A conservative threshold of 3.0 is used to avoid excluding medically important but visually plain content.

\begin{table*}[ht!] 
\centering 
\caption{Benchmarking video generation models on Med-VBench. Within each segment, \textbf{bold} highlights the best scores. The warping error score was derived by aggregating the scores from all three evaluators.}
\resizebox{0.9\textwidth}{!}{ 
\setlength{\tabcolsep}{3pt} 
\begin{tabularx}{\textwidth}{@{}l |>{\columncolor[rgb]{0.87,0.94,1}\centering\arraybackslash}X| *{5}{>{\centering\arraybackslash}X} >{\centering\arraybackslash}X@{}} 
\toprule 
\textbf{Model} & \cellcolor[rgb]{0.87,0.94,1}\textbf{Total $\uparrow$} & \textbf{Imaging Quality $\uparrow$} & \textbf{Subject Consistency $\uparrow$} & \textbf{Background Consistency $\uparrow$} & \textbf{Motion Smoothness $\uparrow$} & \textbf{Warping Error $\downarrow$} \\ 
\midrule 
\multicolumn{7}{l}{\textbf{Open-Source Video Generation Models}} \\ 
\midrule 
CogVideoX-2B & 60.05 & 61.22 & 94.03 & 95.01 & 98.04 & 88.00 \\  
CogVideoX-5B & 60.70 & 56.10 & 93.92 & 96.04 & 98.13 & 80.00 \\  
Wan2.1-T2V-1.3B & 59.59 & 53.81 & 90.48 & 92.75 & 98.03 & 77.50 \\  
OpenSora (v1.2) & 59.44 & 64.88 & 95.40 & 96.46 & 99.40 & 99.50 \\  
OpenSoraPlan (v1.3) & 61.39 & 67.60 & 97.08 & 97.38 & 99.33 & 93.03 \\  
VideoCrafter-2 & 61.43 & 70.52 & \textbf{98.13} & \textbf{98.42} & 98.53 & 97.00 \\ 
ModelScope-1.7B & 58.27 & 63.99 & 92.99 & 96.24 & 96.43 & 100.00 \\  
Latte-1 & 60.60 & 69.87 & 96.41 & 96.75 & 98.09 & 97.50 \\  
Vchitect-2 & 56.50 & 63.02 & 88.07 & 93.50 & 93.89 & 99.50 \\  
Pyramid-Flow & 59.10 & 67.41 & 91.54 & 94.75 & 99.38 & 98.50 \\  
Allegro & 60.98 & 72.22 & 92.96 & 95.16 & 99.01 & 93.50 \\  
Mochi-1-preview & 59.84 & 56.38 & 92.51 & 94.60 & 99.08 & 83.50 \\  
LTX-Video & 59.12 & 61.85 & 97.44 & 95.82 & 99.60 & 100.00 \\  
HunyuanVideo & 67.89 & 65.85 & 91.47 & 95.82 & 99.20 & 45.00 \\  
Wan2.1-T2V-14B & 65.13 & 56.27 & 91.86 & 94.13 & 98.51 & 50.00 \\  
MedGen (Ours) & \textbf{70.93} & \textbf{72.50} & 94.41 & 97.39 & \textbf{99.76} & \textbf{38.50} \\ 
\midrule 
\multicolumn{7}{l}{\textbf{Proprietary Video Generation Models}} \\ 
\midrule 
Pika (v2.0) & 70.29 & 69.62 & \textbf{97.59} & \textbf{97.26} & \textbf{99.57} & 42.29 \\ 
Hailuo (video-01) & 69.45 & 69.84 & 94.55 & 95.24 & 99.35 & 42.27 \\ 
Kling (v1.6) & \textbf{72.32} & \textbf{74.42} & 95.64 & 93.19 & 97.39 & \textbf{26.70} \\ 
Sora & 71.92 & 71.96 & 93.40 & 95.36 & 99.21 & 28.40 \\ 
\bottomrule 
\end{tabularx} 
}
\label{tab:vbench-video-generation-model-performance} 
\end{table*}

\textbf{Technical Quality Filtering}
Low-level artifacts such as compression glitches, frame jitter, and bitrate instability can also degrade model training. We assess these factors using Dover~\cite{wu2023dover}, which provides a technical quality score independent of content semantics. Videos with a Dover score greater than 0 are excluded to ensure temporal and encoding consistency.

\textbf{Joint Filtering}
To further eliminate severely degraded content, we apply a stricter combined filter: videos with both Dover score \textgreater 0.3 and LAION score \textless 4.0 are removed. This approach balances aesthetic and technical quality, helping preserve clinically valuable yet visually simple videos.

\subsection{Data Statistics} \label{sec:data-stats}
\label{sec_33}

\textbf{High-Quality Clips} 
As shown in Figure~\ref{fig:data_distributions}, most video clips in MedVideoCap-55K range from 6 to 10 seconds, striking a balance between semantic richness and training efficiency. This duration is sufficient to capture complete medical actions or concepts, while remaining suitable for current video generation model limits. Each clip is accompanied by both a brief and a detailed caption, supporting a wide range of supervision granularity. In addition, aesthetic and technical quality scores are concentrated in the medium-to-high range, indicating that the dataset maintains a high standard of visual clarity, coherence, and temporal smoothness.

\textbf{Diversity of Medical Domains}
MedVideoCap-55K spans a broad set of medical domains, including clinical procedures, medical education, scientific communication, animations, and imaging, as shown in Figure~\ref{fig:data_distributions}. This diversity supports training models that generalize across various real-world medical contexts. More visual examples and data analysis of MedVideoCap-55K are provided in the Appendix.

\section{Experiments}
\label{sec:training}

In this section, we develop a \textit{medically-adapted} video generation model, \textbf{MedGen}, which is further trained from a \textit{general} video generation model using \textit{MedVideoCap-55K}.

\subsection{Medical Video Generation Model: MedGen}

\begin{figure*}[ht!]
    \centering
    \includegraphics[width=\textwidth]{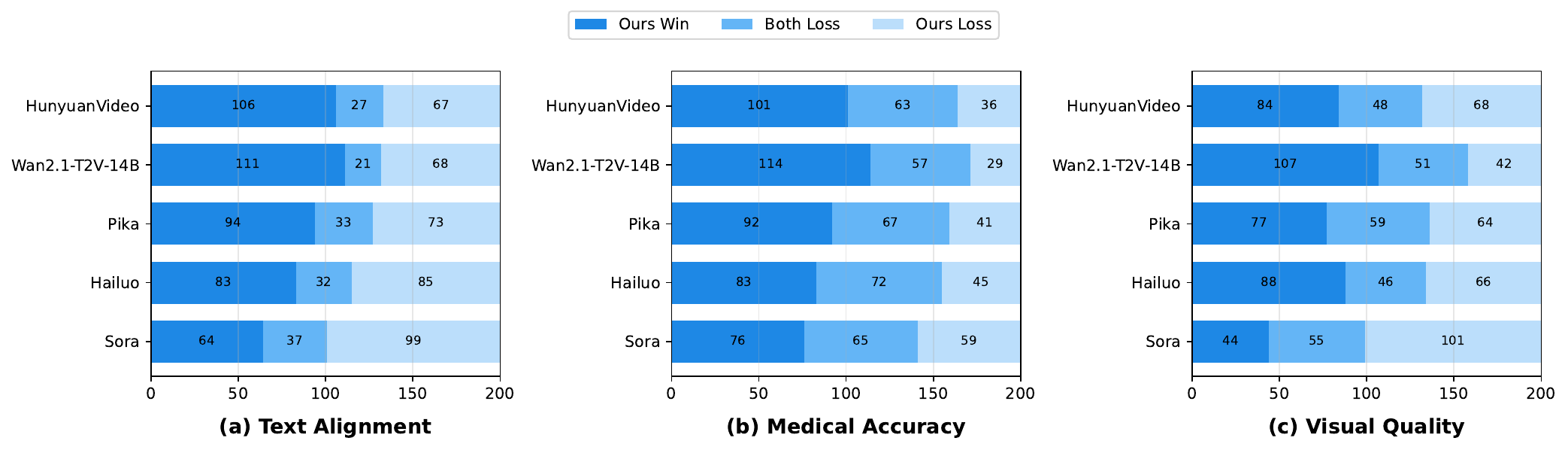}
    \caption{Results of the doctor evaluators assessing the medical videos generated by MedGen and other models across three dimensions. 
    (a) Text Alignment focuses on the consistency between the generated video and the prompt. (b) Medical Accuracy focuses on the generated video adheres to medical common sense. (c) Visual Quality focuses on the overall quality of the generated video.
    }
    \label{fig:human_eval_res}
\end{figure*}

\begin{table*}[ht!]
\centering
\caption{Comparison of data augmentation performance between MedGen and HunyuanVideo in downstream supervised Tasks.}
\label{tab:empirical-study}
\resizebox{0.99\textwidth}{!}{
\begin{tabular}{l|llll|llll|llll}
\toprule
& \multicolumn{4}{c}{MedVidCL} & \multicolumn{4}{c}{HyperKvasir} & \multicolumn{4}{c}{SurgVisDom} 
\\
\cmidrule(lr){2-5} \cmidrule(lr){6-9} \cmidrule(lr){10-13} 
& ACC. & Pre. & Rec. & F1. & ACC. & Pre. & Rec. & F1. & ACC. & Pre. & Rec. & F1. \\
\midrule
$D_{ori}$ & $61.73$ & $61.29$ & $61.12$ &  $61.01$ & $44.44$ & $11.11$ & $25.00$ & $15.38$ & $14.29$ & $5.56$ &  $16.67$ &  $8.33$\\
$D_{ori}$ + $D_{hunyuan}$ & $63.58_{\textcolor{deepgreen}{+1.85}}$ & $63.40_{\textcolor{deepgreen}{+2.11}}$ & $63.25_{\textcolor{deepgreen}{+2.13}}$ & $63.27_{\textcolor{deepgreen}{+2.26}}$ & $44.44_{\textcolor{red}{+0.00}}$ & $21.25_{\textcolor{deepgreen}{+10.14}}$ & $31.25_{\textcolor{deepgreen}{+6.25}}$ & $25.00_{\textcolor{deepgreen}{+9.62}}$ & $28.57_{\textcolor{deepgreen}{+14.28}}$ & $9.52_{\textcolor{deepgreen}{+3.96}}$ &  $33.33_{\textcolor{deepgreen}{+16.66}}$ &  $14.81_{\textcolor{deepgreen}{+6.48}}$\\
\midrule
$D_{ori}$ + $D_{medgen}$ & $64.81_{\textcolor{deepgreen}{+3.08}}$ & $65.61_{\textcolor{deepgreen}{+4.32}}$ & $65.33_{\textcolor{deepgreen}{+4.21}}$ & $64.75_{\textcolor{deepgreen}{+3.74}}$ & $55.56_{\textcolor{deepgreen}{+11.12}}$ & $26.79_{\textcolor{deepgreen}{+15.68}}$ & $37.50_{\textcolor{deepgreen}{+12.50}}$ & $30.68_{\textcolor{deepgreen}{+15.30}}$ & $42.86_{\textcolor{deepgreen}{+28.57}}$ & $14.29_{\textcolor{deepgreen}{+8.73}}$ &  $33.33_{\textcolor{deepgreen}{+16.66}}$ &  $20.00_{\textcolor{deepgreen}{+11.67}}$\\
\bottomrule
\end{tabular}
}
\end{table*}

\textbf{Experimental Setup}  
MedGen is built on top of the open-source latent diffusion video model HunyuanVideo\footnote{\url{https://huggingface.co/tencent/HunyuanVideo}}. We train MedGen on 8 NVIDIA A800 GPUs using 50,000 steps with a batch size of 32, LoRA rank set to 32, and a learning rate of 5e-5. Each input sample consists of a medical video clip and a corresponding detailed caption from MedVideoCap-55K. Training focuses on improving the model’s ability to generate semantically aligned and medically accurate videos from text prompts. Further training details are provided in Appendix. 

\textbf{Baselines}
We compare MedGen with 15 open-source and 4 commercial state-of-the-art video generation models. The open-source baselines include popular latent diffusion and autoregressive models such as CogVideoX~\cite{yang2024cogvideox}, Mochi-1-preview~\cite{genmo2024mochi}, and Wan2.1~\cite{wan2.1}, while the commercial models include Sora~\cite{sora}, Pika~\cite{pika}, and Hailuo~\cite{hailuo}.

\textbf{Evaluation Metrics}
We evaluate all the models with the following metrics. \textbf{(1) Med-VBench}. VBench~\cite{huang2024vbench} is a widely used benchmark for evaluating general video generation models, breaking down video quality into specific, hierarchical, and independent dimensions with tailored metrics. Building on this framework, we introduce medical video prompts and rename the benchmark Med-VBench to reflect its domain-specific focus. As aesthetic quality is less relevant for medical videos, we omit this dimension from our evaluation. \textbf{(2) Human Evaluation}. Besides the automatic metrics, we also use human evaluation, we invited three doctors to score these models. Specifically, for medical videos generated by different models based on the same prompt, the experts assessed them across three dimensions: Text Alignment, Medical Accuracy, and Visual Quality. More details on evaluation design and implementation are provided in Appendix.

\subsection{Main Results}

\textbf{Automatic Evaluation Results} Table~\ref{tab:vbench-video-generation-model-performance} compares \textit{MedGen} against a range of open-source and commercial video generation models using the Med-VBench benchmark. \textbf{MedGen achieves the best performance among all open-source models}. Notably, MedGen excels in \textit{Factual Consistency}, \textit{Text-to-Video Alignment}, and \textit{Imaging Quality}, while maintaining one of the lowest warping error scores across all models. These results highlight MedGen’s strong ability to generate medically accurate and visually coherent content. Commercial systems like Sora, Pika, Kling, and Hailuo outperform most open-source models, indicating a general quality gap likely due to access to larger and more diverse training data. However, MedGen narrows this gap significantly, especially on medically relevant metrics. Interestingly, model size does not always correlate with performance in the medical domain. For example, \textit{Wan2.1-T2V-1.3B} achieves similar results to \textit{CogVideoX-5B}, suggesting that domain adaptation and training data quality are more critical than parameter count. In summary, MedGen offers a strong open-source alternative for medical video generation, combining domain precision with competitive visual quality.

\textbf{Human Evaluation Results}
Figure~\ref{fig:human_eval_res} shows the results of human evaluation. Compared with most video generation models, MedGen performs excellently across three dimensions: Text Alignment, Medical Accuracy, and Visual Quality. Its outstanding performance in Medical Accuracy further validates the effectiveness of the MedVideoCap-55K. In addition, the Cohen’s Kappa coefficient among the three independent experts was above 0.75, indicating strong inter-rater agreement and confirming the reliability of the scoring process. More details on Cohen’s Kappa coefficient can be found in Appendix.

\subsection{Transferability of MedVideoCap-55K on Mochi}

\begin{table}[ht!]
\centering
\caption{Performance comparison of original vs. MedVideoCap-55K-trained Mochi-1-preview on VBench. Metrics (abbreviated) follow Table \ref{tab:vbench-video-generation-model-performance}. "Original" is the base model; "Trained" is fine-tuned on MedVideoCap-55K. $^\dagger$ indicates statistically significant improvements based on a two-tailed t-test (significance level $p$ = 0.05).}
\resizebox{\columnwidth}{!}{
\begin{tabular}{@{}l | l | *{5}{l}@{}}
\toprule
\textbf{} & \cellcolor[rgb]{0.87,0.94,1}\textbf{Total $\uparrow$} & \textbf{WE $\downarrow$} & \textbf{IQ $\uparrow$} & \textbf{SC $\uparrow$} & \textbf{BC $\uparrow$} & \textbf{MS $\uparrow$} \\
\midrule
Original & \cellcolor[rgb]{0.87,0.94,1}59.84 & 83.50 & 56.38 & 92.51 & \textbf{94.60} & \textbf{99.08} \\
\midrule
Trained & \cellcolor[rgb]{0.87,0.94,1}$62.77_{\textcolor{deepgreen}{+2.93}}$ & \textbf{72.00}$^\dagger$ & \textbf{63.42}$^\dagger$ & \textbf{94.27}$^\dagger$ & 93.11 & 97.82 \\
\bottomrule
\end{tabular}
}
\label{tab:vbench-mochi}
\end{table}

\begin{figure*}[ht]
    \centering
    \includegraphics[width=0.93\textwidth]{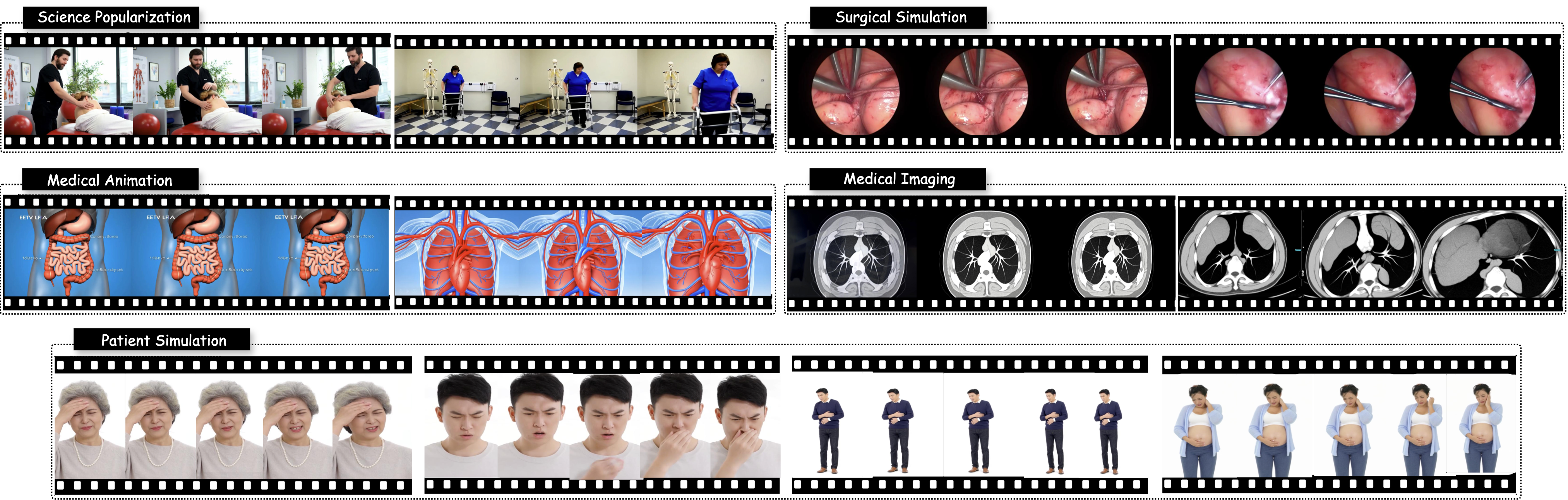}
    \caption{Examples of MedGen-generated videos across diverse real-world medical scenarios.}
    \label{fig:case-show}
\end{figure*}

\textbf{Experimental Setup.} To verify the transferability of our constructed MedVideoCap-55K, we switched the base model from HunyuanVideo  to the Mochi-1-preview model~\cite{genmo2024mochi}. We conducted LoRA training for Mochi-1-preview~\footnote{\url{https://huggingface.co/genmo/mochi-1-preview}} using MedVideoCap-55K. The training setup remained consistent with the configurations used in MedGen training.

\textbf{Results} Table~\ref{tab:vbench-mochi} shows a performance comparison between the original Mochi-1-preview model and the further-trained Mochi-1-preview model using MedVideoCap-55K. Experiments demonstrates that training on our MedVideoCap-55K can  significantly enhance the medical video generation capabilities in metrics like warping wrror, image quality and  subject consistency; this confirms the quality of our dataset MedVideoCap-55K. The slight decline in background consistency and motion smoothness may stem from the limited diversity of the training dataset. We hypothesize that incorporating a balanced proportion of non-medical video data could mitigate this issue while preserving performance in these metrics. Exploring this approach to enhance dataset variety and its impact on model robustness remains an avenue for future work.

\section{Applications of MedGen}

\subsection{Application I: Data Augmentation}
We evaluate MedGen's effectiveness as a data augmenter for downstream medical video analysis. As shown in Table~\ref{tab:empirical-study}, we use MedGen-generated videos to augment training data across three classification benchmarks: MedVidCL \cite{gupta2023dataset}, HyperKvasir \cite{borgli2020hyperkvasir}, and SurgVisDom \cite{schoeffmann2018video}. Compared to using only the original annotated data ($D_{ori}$) and augmentation with HunyuanVideo, incorporating MedGen-generated data leads to consistently higher gains across all metrics and datasets. Notably, MedGen improves F1 scores by up to \textbf{+15.3} on HyperKvasir and \textbf{+11.7} on SurgVisDom, outperforming HunyuanVideo by a clear margin. These results demonstrate MedGen’s strong potential as a high-quality, domain-relevant data source for enhancing medical video understanding tasks. 

\subsection{Application II: Science Popularization and User Simulation}

MedGen shows strong potential across a wide range of medical video generation scenarios, including surgical training, patient education, medical animation, and remote consultation. As illustrated in Figure~\ref{fig:case-show}, MedGen can generate diverse and high-quality videos spanning science popularization, surgical simulation, medical imaging, educational content, and patient simulations. This versatility enables MedGen to serve as a powerful tool for \textbf{Science Popularization and User Simulation}, particularly in scenarios where real video data is scarce, privacy-sensitive, or costly to obtain. Its ability to produce visually coherent and medically relevant content makes it well-suited for augmenting datasets, prototyping simulations, or enhancing medical communication. 

\section{Related Work}

\textbf{Text-to-Video Generation.} Video generation has made significant progress over the past two years. The introduction of Sora~\cite{sora} has sparked significant research interest in text-to-video generation. Commercial models such as Kling~\cite{kling}, and Hailuo~\cite{hailuo} demonstrate strong performance. Meanwhile, open-source models such as LTX-Video~\cite{hacohen2024ltx}, CogVideoX~\cite{yang2024cogvideox} and Mochi-1~\cite{genmo2024mochi} enable researchers to experiment with, customize, and enhance existing frameworks. Despite their power, general video generation models face limitations in medical applications due to the scarcity of medical video data, which hinders the professionalism and practicality of generated content.

\textbf{Generation Models in Medical Domain.} 
Video generation has shown great potential in the medical field, with applications including medical concept explanation, disease simulation, and biomedical data augmentation~\cite{li2024artificialintelligencebiomedicalvideo}. In the field of endoscopy and surgery, Endora~\cite{li2024endora} is a generative model designed as an endoscopy simulator, capable of replicating diverse endoscopic scenarios for educational purposes.  
MedSora~\cite{wang2024optical} combines spatio-temporal Mamba modules, optical flow alignment, and frequency-compensated video VAEs to generate high-quality medical videos, improving performance on downstream classification tasks. 
Additionally, Surgen~\cite{cho2024surgentextguideddiffusionmodel} and SurgSora~\cite{chen2024surgsora} can generate highly realistic surgical operation videos based on text instructions, opening up new possibilities for surgical simulation. Unlike previous works, our goal is to develop  medical video generation models that can meet a wide range of medical and healthcare needs.

\textbf{Text-to-Video Datasets.} Data quality and quantity are closely related to model performance. For general domains, many text-video datasets, such as OpenVid~\cite{nan2024openvid}, Panda~\cite{wang2020pandagigapixellevelhumancentricvideo}, and WebVid~\cite{bain2022frozentimejointvideo}, have been proposed to advance multimodal understanding and generation. However, despite their extensive and diverse collections of text-video pairs, these datasets are not directly applicable to the medical domain. Additionally, there are several datasets related to medical videos, such as MedVidCL~\cite{gupta2023dataset}, Colonoscopy~\cite{mesejo2016computer}, Kvasir-Capsule~\cite{borgli2020hyperkvasir}, and CholecTriplet~\cite{nwoye2022rendezvous}, which are often utilized for performing supervised classification tasks.

\section{Conclusion}
Current models lack clinical accuracy in medical video generation, largely due to the scarcity of high-quality annotated datasets. To this end, we introduce \textbf{MedVideoCap-55K}, a large-scale, diverse, and caption-rich medical video dataset designed to address the lack of high-quality data for medical video generation. With over 55,000 curated clips spanning a broad range of real-world medical scenarios, MedVideoCap-55K provides the foundation for training generalist medical video generation models. Built on it, our model \textbf{MedGen} achieves state-of-the-art performance among open-source models and rivals commercial systems across multiple benchmarks. In addition, we demonstrate its utility for data augmentation in downstream medical video analysis tasks, showing significant performance gains. Our findings highlight the importance of domain-specific datasets and tailored evaluation in advancing medical video generation. We hope MedVideoCap-55K and MedGen will provide a valuable reference for the field of medical video generation, while also driving further advancements and innovation in related research.

\textbf{Ethics Statement}~
All data are publicly available, compliant with YouTube’s terms, and exclude personal/sensitive content. Captions were auto-generated (MLLMs) and manually verified to remove inappropriate/identifiable material. The dataset is intended solely and strictly for research purposes and should not be used for non-research settings, especially in clinical practice.

\section*{Acknowledgment}

This work was supported by the Shenzhen Science and Technology Program (JCYJ20220818103001002), Shenzhen Doctoral Startup Funding (RCBS20221008093330065), Tianyuan Fund for Mathematics of National Natural Science Foundation of China (NSFC) (12326608), Shenzhen Science and Technology Program (Shenzhen Key Laboratory Grant No. ZDSYS20230626091302006), and Shenzhen Stability Science Program 2023, Shenzhen Key Lab of Multi-Modal Cognitive Computing.



\bibliography{aaai2026}

\clearpage
\appendix
\onecolumn




\section{Experiments}

\subsection{Training Details on MedGen}
\label{app:training-details}

\begin{figure}[htbp]
\centering
\noindent
\begin{minipage}[t]{0.40\textwidth}
\centering
\vspace{0pt} 
\begin{tabular}{lc}
\hline
\textbf{Argument} & \textbf{Value} \\
\hline
$\texttt{--seed}$ & 1024 \\
$\texttt{--train\_batch\_size}$ & 32 \\
$\texttt{--dataloader\_num\_workers}$ & 4 \\
$\texttt{--gradient\_accumulation\_steps}$ & 8 \\
$\texttt{--learning\_rate}$ & 5e-5 \\
$\texttt{--mixed\_precision}$ & bf16 \\
$\texttt{--checkpointing\_steps}$ & 100 \\
$\texttt{--validation\_steps}$ & 1000 \\
$\texttt{--validation\_sampling\_steps}$ & 50 \\
$\texttt{--checkpoints\_total\_limit}$ & 100 \\
$\texttt{--allow\_tf32}$ & True \\
$\texttt{--ema\_start\_step}$ & 0 \\
$\texttt{--cfg}$ & 0.0 \\
$\texttt{--ema\_decay}$ & 0.999 \\
$\texttt{--num\_frames}$ & 93 \\
$\texttt{--validation\_guidance\_scale}$ & "1.0" \\
$\texttt{--shift}$ & 7 \\
$\texttt{--use\_lora}$ & True \\
$\texttt{--lora\_rank}$ & 32 \\
$\texttt{--lora\_alpha}$ & 32 \\
\hline
\end{tabular}
\captionof{table}{Training configuration.}
\label{tab:train-settings}
\end{minipage}
\hfill
\begin{minipage}[t]{0.48\textwidth}
\centering
\vspace{0pt} 
\includegraphics[width=0.65\textwidth]{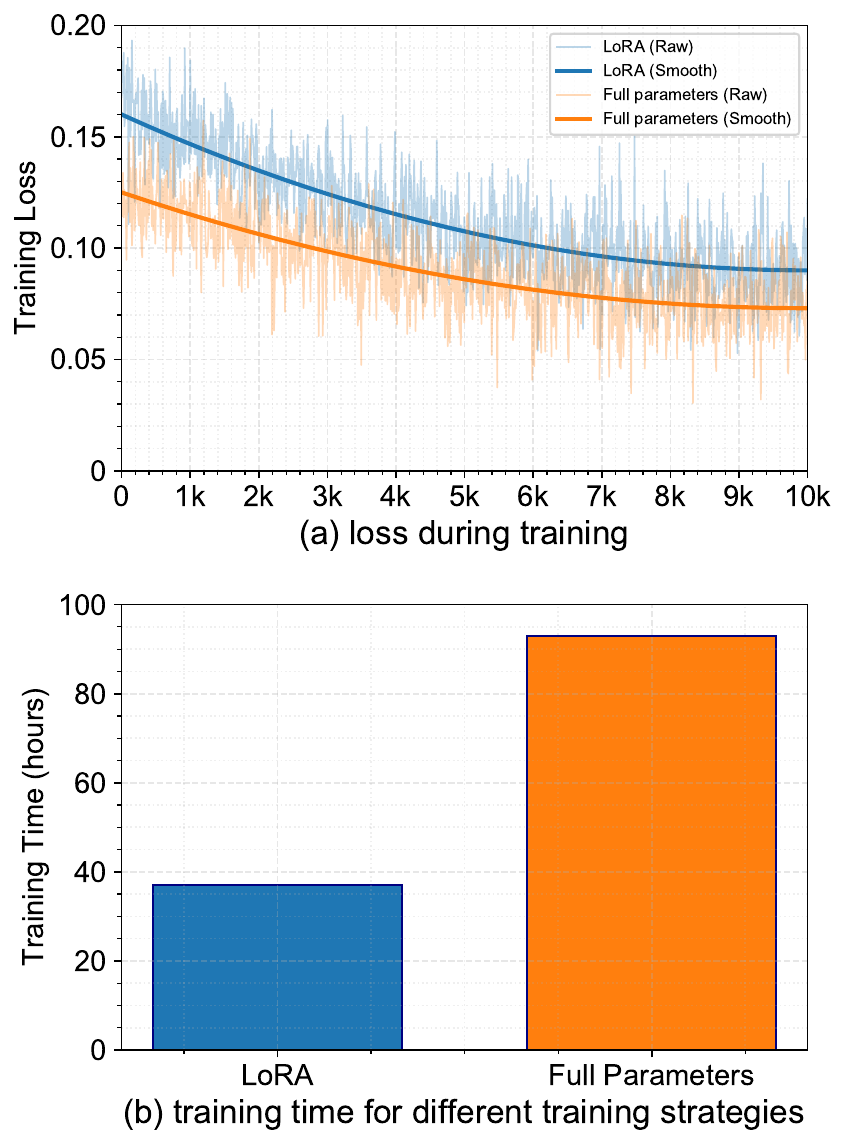}
\caption{Comparison of training efficiency. (a): the change of loss during training. (b): training time for different training strategies.}
\label{fig:train-eff-comp}
\end{minipage}
\end{figure}

Table~\ref{tab:train-settings} shows the detailed training parameter settings used to train MedGen.

We compared the differences in training efficiency between using LoRA and full-parameter training. Figure~\ref{fig:train-eff-comp} (a) shows the variation of the loss during training with the same number of steps, comparing the use of LoRA and full-parameter fine-tuning. Figure~\ref{fig:train-eff-comp} (b) demonstrates that, under the same number of training steps, LoRA required only half the GPU runtime to update its internal parameters.


\subsection{Baseline Models}
\label{app:baseline-model}

\begin{table}[ht]
\centering
\resizebox{0.8\columnwidth}{!}{
\begin{tabular}{l|l}
\hline
\textbf{Model} & \textbf{Model Link} \\
\hline
\multicolumn{2}{c}{\textbf{Open-Source Video Generation Model}} \\
\hline
CogVideoX-2B~\cite{yang2024cogvideox} & \url{https://huggingface.co/THUDM/CogVideoX-2b} \\
CogVideoX-5B~\cite{yang2024cogvideox} & \url{https://huggingface.co/THUDM/CogVideoX-5b} \\
Wan2.1-T2V-1.3B~\cite{wan2.1} & \url{https://huggingface.co/Wan-AI/Wan2.1-T2V-1.3B} \\
OpenSora V1.2~\cite{zheng2024open} & \url{https://huggingface.co/hpcai-tech/Open-Sora-v2} \\
OpenSoraPlan V1.3~\cite{lin2024open} & \url{https://huggingface.co/LanguageBind/Open-Sora-Plan-v1.3.0} \\
VideoCrafter-2~\cite{chen2024videocrafter2} & \url{https://huggingface.co/VideoCrafter/VideoCrafter2} \\
ModelScope-1.7B~\cite{wang2023modelscope} & \url{https://huggingface.co/ali-vilab/text-to-video-ms-1.7b} \\
Latte-1~\cite{ma2024latte} & \url{https://huggingface.co/maxin-cn/Latte-1} \\
Vchitect-2~\cite{fan2025vchitect} & \url{https://huggingface.co/Vchitect/Vchitect-2.0-2B} \\
Pyramid-Flow~\cite{jin2024pyramidalflowmatchingefficient} & \url{https://huggingface.co/rain1011/pyramid-flow-sd3} \\
Allegro~\cite{allegro2024} & \url{https://huggingface.co/rhymes-ai/Allegro} \\
Mochi-1-preview~\cite{genmo2024mochi} & \url{https://huggingface.co/genmo/mochi-1-preview} \\
LTX-Video~\cite{hacohen2024ltx} & \url{https://huggingface.co/Lightricks/LTX-Video} \\
HunyuanVideo~\cite{kong2024hunyuanvideo} & \url{https://huggingface.co/tencent/HunyuanVideo} \\
Wan2.1-T2V-14B~\cite{wan2.1} & \url{https://huggingface.co/Wan-AI/Wan2.1-T2V-14B} \\
\hline
\multicolumn{2}{c}{\textbf{Commercial Video Generation Model}} \\
\hline
Sora~\cite{sora} & \url{https://sora.com/} \\
Pika V2.0~\cite{pika} & \url{https://pika.art/} \\
Hailuo~\cite{hailuo} & \url{https://hailuoai.video/} \\
Kling V1.6 & \url{https://klingai.com/} \\
\hline
\end{tabular}
}
\caption{Baseline Video Generation Models and Their Official Links.}
\label{tab:video-models}
\end{table}

\begin{table}[ht]
\centering
\caption{Deformation Evaluation Criteria.}
\label{tab:de-evaluation}
\begin{tabularx}{\columnwidth}{@{}lX@{}}
\toprule
\textbf{Label} & \textbf{Evaluation Criteria} \\ \midrule

No Distortion  & No deformation was observed, and the image is fully intact. \\

\midrule
Minor Distortion & The distortion is not significant or is only a slight local deformation. Overall, the original form of the characters, objects, or scenes can still be recognized, and viewers will not experience noticeable discomfort or disturbance. \\

\midrule
Moderate Distortion & The shape of the characters or objects may deviate significantly, and certain details in the image may become blurred or distorted. The distortion may affect the recognizability of some scenes, but it will not disappear entirely. \\

\midrule
Severe Distortion & The shape of the characters or scenes undergoes extreme distortion, with effects such as unrecognizable twisting, compression, stretching, or even complete distortion to the point of being unidentifiable. \\

\bottomrule
\end{tabularx}
\end{table}

\begin{figure}[ht]
    \centering
    \includegraphics[width=0.70\textwidth]{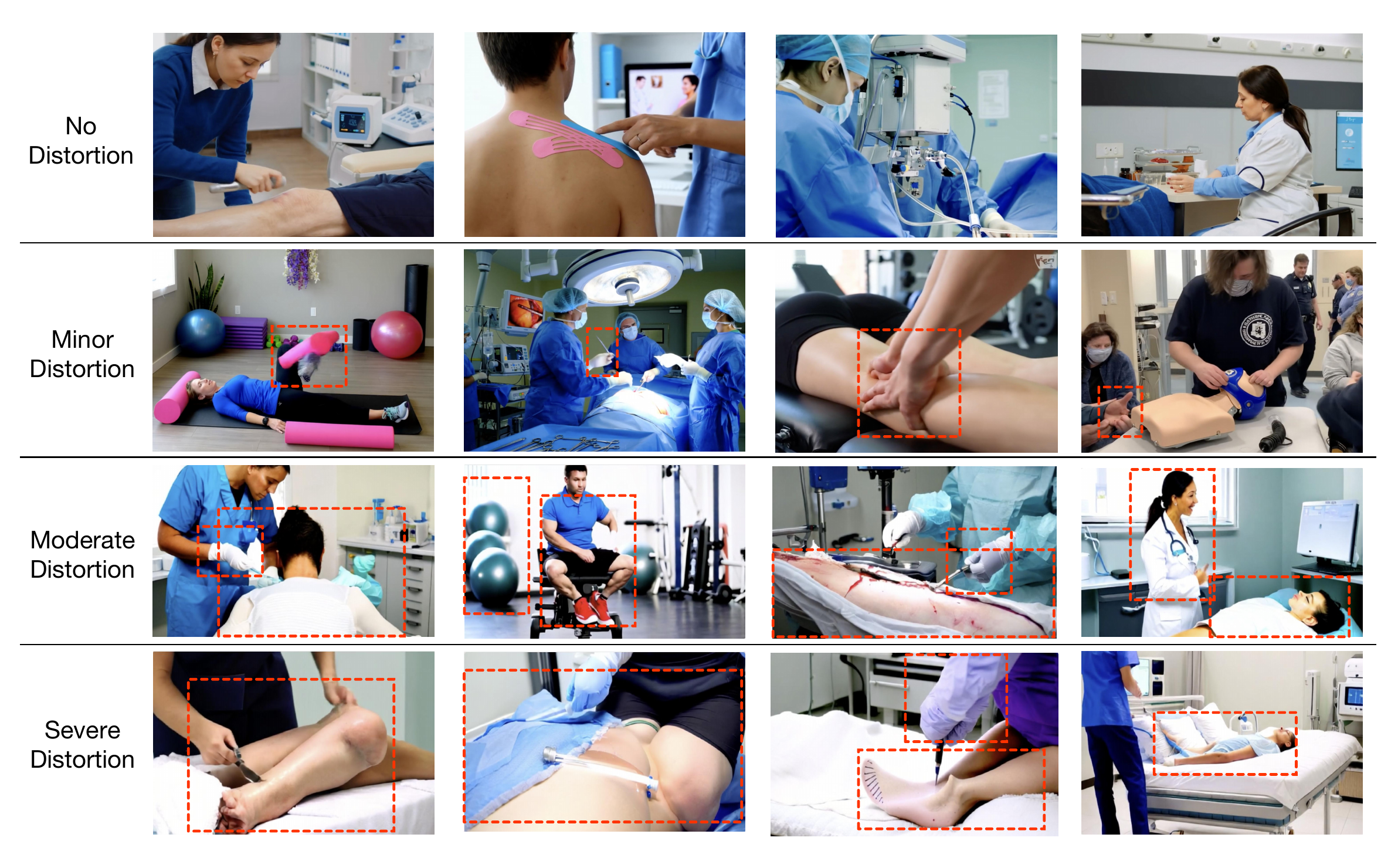}
    \caption{Example video clips of distortion at different levels, with the distorted areas highlighted in red boxes.}
    \label{fig:Distortion-example-figure}
\end{figure}

All tests for commercial video generation models were conducted using the latest versions available before \underline{February 10, 2025}. Pika used version 2.0, Kling used version 1.6, while Hailuo adopted the video-01 version.

\subsection{Evaluation Details}
\label{app:eval-details}

To prompt all models to generate medical videos for evaluation, we designed 200 different prompts, covering a variety of medical video categories and medical scenarios. For fair comparison, we conducted inference only once and maintained the default settings for all selected models, avoiding any cherry-picking of results. Additionally, we found that although many videos generated by the models received high scores, they suffered from severe distortion, which could significantly affect the accurate delivery of information. To address this issue, we invited three human evaluators to specifically assess the degree of distortion in the videos, and the warping error score was derived by aggregating the scores from all three evaluators.

We predefined evaluation criteria for different levels of image distortion, as illustrated in Figure~\ref{fig:Distortion-example-figure}, and detailed the corresponding assessment guidelines in Table~\ref{tab:de-evaluation}. Three human evaluators were invited to independently assess each video and assign a distortion level accordingly. If at least two of the evaluators rated a video as “No Distortion” or “Minor Distortion,” the medical video was considered acceptable. Conversely, if two of the evaluators rated it as “Moderate Distortion” or “Severe Distortion,” the video was deemed to exhibit an unacceptable level of distortion.

To evaluate the similarity in scoring among the three evaluators, we employed Cohen's Kappa coefficient to quantify the inter-annotator agreement. This statistical method is widely used to assess the degree of consistency in classification tasks, especially when multiple evaluators make judgments under the same conditions. The results showed that the agreement score between Evaluator 1 and Evaluator 2 was 0.82, between Evaluator 1 and Evaluator 3 was 0.79, and between Evaluator 2 and Evaluator 3 was 0.77. Overall, these relatively high Kappa values indicate a strong level of consistency and scoring reliability among the three evaluators in the video distortion assessment task.

\begin{mybox}[colback=gray!10]{One of 200 Prompts used for Evaluation}

The short video depicts a surgical procedure taking place in an operating room. The scene is filled with medical professionals, all dressed in surgical attire, including scrubs, masks, and caps. They are gathered around a patient who is lying on an operating table, covered with sterile drapes. The focus is on the surgical team as they perform the operation, with various medical instruments and equipment visible in the background. The lighting is concentrated on the surgical area, providing a clear view of the procedure being conducted. The team appears to be working collaboratively, with some members actively engaged in the surgery while others assist or monitor the process. The overall atmosphere is one of precision and care, characteristic of a professional medical setting.

\end{mybox}

To evaluate MedGen's performance in practical applications, we invited three doctors to participate in the assessment. The specific method was as follows: we randomly selected a model to compare with MedGen and presented the videos generated by both models—based on the same text prompts—in an anonymous manner to the doctors. The doctors judged the videos from three perspectives: Text Alignment, Medical Accuracy, and Visual Quality. For each dimension, the doctors were asked to choose the better-performing model. If they believed neither performed satisfactorily, they could select "Both Loss".

\subsection{VideoScore Evaluation Result}

In addition to the experiments on Med-VBench, we conducted supplementary evaluations on the VideoScore to comprehensively assess the generalization and robustness of our model, MedGen. VideoScore~\cite{he2024videoscore} is an end-to-end video generation evaluation benchmark whose results exhibit a high correlation with human judgment. We used version 1.1 of VideoScore. These additional experiments further validate MedGen's competitiveness in mainstream video generation tasks, highlighting its strong modeling capabilities. The results are shown in Figure~\ref{tab:videoscore-video-generation-model-performance}.

\begin{table*}[ht!]
\centering
\caption{Benchmarking video generation models on VideoScore. Within each segment, \textbf{bold} highlights the best scores. The warping error score was derived by aggregating the scores from all three evaluators.}
\resizebox{0.8\textwidth}{!}{
\setlength{\tabcolsep}{2.5pt}
\begin{tabularx}{\textwidth}{@{}l |>{\columncolor[rgb]{0.87,0.94,1}\centering\arraybackslash}X| *{6}{>{\centering\arraybackslash}X}@{}}
\toprule
Model & \cellcolor[rgb]{0.87,0.94,1}Total $\uparrow$ & Visual Quality $\uparrow$ & Temporal Consistency $\uparrow$ & Text-to-Video Alignment $\uparrow$ & Factual Consistency $\uparrow$ & Warping Error $\downarrow$ \\
\midrule
\multicolumn{7}{l}{Open-Source Video Generation Models} \\
\midrule
CogVideoX-2B & 2.21 & 3.35 & 3.21 & \textbf{3.13} & 3.09 & 88.00 \\ 
CogVideoX-5B & 2.22 & 3.30 & 3.13 & 3.07 & 3.03 & 80.00 \\ 
Wan2.1-T2V-1.3B & 2.12 & 3.11 & 2.89 & 2.99 & 2.86 & 77.50 \\
OpenSora (v1.2) & 2.03 & 3.21 & 3.04 & 2.97 & 2.93 & 99.50 \\ 
OpenSoraPlan (v1.3) & 2.21 & \textbf{3.39} & 3.29 & \textbf{3.13} & 3.18 & 93.03 \\ 
VideoCrafter-2 & 1.88 & 2.91 & 2.78 & 2.80 & 2.64 & 97.00 \\
ModelScope-1.7B & 1.82 & 2.86 & 2.72 & 2.77 & 2.58 & 100.00 \\ 
Latte-1 & 1.95 & 3.08 & 2.87 & 2.91 & 2.76 & 97.50 \\ 
Vchitect-2 & 1.88 & 2.94 & 2.80 & 2.82 & 2.67 & 99.50 \\
Pyramid-Flow & 1.84 & 2.87 & 2.74 & 2.76 & 2.60 & 98.50 \\
Allegro & 2.09 & 3.28 & 3.12 & 2.82 & 3.06 & 93.50 \\
Mochi-1-preview & 2.02 & 3.03 & 2.84 & 2.84 & 2.73 & 83.50 \\
LTX-Video & 1.90 & 3.01 & 2.83 & 2.83 & 2.71 & 100.00 \\
HunyuanVideo & 2.35 & 3.16 & 3.01 & 2.82 & 2.91 & 45.00 \\ 
Wan2.1-T2V-14B & 2.25 & 2.99 & 2.77 & 2.93 & 2.68 & 50.00 \\
MedGen (Ours) & \textbf{2.57} & 3.35 & \textbf{3.30} & 3.09 & \textbf{3.22} & \textbf{38.50} \\
\midrule
\multicolumn{7}{l}{Proprietary Video Generation Models} \\
\midrule
Pika (v2.0) & 2.19 & 2.80 & 2.85 & 2.53 & 2.63 & 42.29  \\
Hailuo (video-01) & 2.27 & 2.99 & 2.94 & 2.63 & 2.78 & 42.27 \\
Kling (v1.6) & 2.80 & 3.55 & 3.56 & \textbf{3.42} & \textbf{3.86} & \textbf{26.70} \\
Sora & \textbf{2.96} & \textbf{3.97} & \textbf{3.85} & 3.24 & 3.82 & 28.40 \\
\bottomrule 
\end{tabularx}
}
\label{tab:videoscore-video-generation-model-performance}
\end{table*}

\subsection{Inter-Rater Reliability Among Expert Evaluators}
\label{app:expert-cohen}

We used Cohen’s Kappa coefficient to measure agreement among the three medical experts. Table~\ref{tab:cohen_kappa} indicate strong agreement among the experts, confirming the reliability of the scoring process.

\begin{table}[ht]
\centering
\begin{tabular}{l|c}
\hline
\textbf{Comparison Pair} & \textbf{Cohen’s Kappa} \\
\hline
Expert1 vs. Expert2 & 0.83 \\
Expert1 vs. Expert3 & 0.81 \\
Expert2 vs. Expert3 & 0.78 \\
\hline
\end{tabular}
\caption{Comparison of Cohen’s Kappa values between experts.}
\label{tab:cohen_kappa}
\end{table}

\subsection{The generalizability of MedGen}

To evaluate MedGen's generalization ability on unseen video types, we trained the HunyuanVideo model using a subset of 10k medical teaching videos extracted from MedVideoCap-55K. Subsequently, we evaluated its performance on various unseen medical video categories using the VBench. The results are summarized in the Table~\ref{tab:gen_vbench_results}.

\begin{table*}[t]
\centering
\resizebox{0.75\textwidth}{!}{%
\begin{tabular}{l|c|c|c|c}
\hline
\textbf{Model} & \textbf{Clinical Practice} & \textbf{Animation} & \textbf{Popular Science} & \textbf{Imaging} \\
\hline
HunyuanVideo & 62.53 & 65.19 & 72.33 & 55.17 \\
+ trained on 10k teaching videos & 64.30 (+1.77) & 62.11 (-3.08) & 75.01 (+2.68) & 58.15 (+2.98) \\
\hline
\end{tabular}%
}
\caption{Med-VBench results of HunyuanVideo on unseen video types.}
\label{tab:gen_vbench_results}
\end{table*}

The results indicate that even when trained on unrelated datasets, the model can still demonstrate strong generalization performance on unseen data.

\end{document}